\documentclass[default,iicol,sn-mathphys-num]{sn-jnl}

\usepackage{graphicx}%
\usepackage{multirow}%
\usepackage{amsmath,amssymb,amsfonts}%
\usepackage{amsthm}%
\usepackage{mathrsfs}%
\usepackage[title]{appendix}%
\usepackage{xcolor}%
\usepackage{textcomp}%
\usepackage{manyfoot}%
\usepackage{booktabs}%
\usepackage{algorithm}%
\usepackage{algorithmicx}%
\usepackage{algpseudocode}%
\usepackage{listings}%

\begin{document}

\title[Article Title]{Real-Time Dynamic Scale-Aware Fusion Detection Network: Take Road Damage Detection as an example}


\author*[1]{\fnm{Weichao} \sur{Pan}}\email{202211107025@stu.sdjzu.edu.cn}

\author[1]{\fnm{Xu} \sur{Wang}}\email{202311102025@stu.sdjzu.edu.cn}

\author[2]{\fnm{Wenqing} \sur{Huan}}\email{202212105033@stu.sdjzu.edu.cn}

\affil*[1]{\orgdiv{School of Computer Science and Technology}, \orgname{Shandong Jianzhu University}, \orgaddress{\street{Ganggou Street}, \city{Jinan}, \postcode{250000}, \state{Shandong}, \country{China}}}

\affil[2]{\orgdiv{School of Science}, \orgname{Shandong Jianzhu University}, \orgaddress{\street{Ganggou Street}, \city{Jinan}, \postcode{250000}, \state{Shandong}, \country{China}}}


\abstract{Unmanned Aerial Vehicle (UAV)-based Road Damage Detection (RDD) is important for daily maintenance and safety in cities, especially in terms of significantly reducing labor costs. However, current UAV-based RDD research is still faces many challenges. For example, the damage with irregular size and direction, the masking of damage by the background, and the difficulty of distinguishing damage from the background significantly affect the ability of UAV to detect road damage in daily inspection. To solve these problems and improve the performance of UAV in real-time road damage detection, we design and propose three corresponding modules: a feature extraction module that flexibly adapts to shape and background; a module that fuses multiscale perception and adapts to shape and background ; an efficient downsampling module. Based on these modules, we designed a multi-scale, adaptive road damage detection model with the ability to automatically remove background interference, called Dynamic Scale-Aware Fusion Detection Model (RT-DSAFDet). Experimental results on the UAV-PDD2023 public dataset show that our model RT-DSAFDet achieves a mAP50 of 54.2\%, which is 11.1\% higher than that of YOLOv10-m, an efficient variant of the latest real-time object detection model YOLOv10, while the amount of parameters is reduced to 1.8M and FLOPs to 4.6G, with a decreased by 88\% and 93\%, respectively. Furthermore, on the large generalized object detection public dataset MS COCO2017 also shows the superiority of our model with mAP50-95 is the same as YOLOv9-t, but with 0.5\% higher mAP50, 10\% less parameters volume, and 40\% less FLOPs.}




\maketitle

\section{Introduction}\label{sec1}

Road Damage Detection (RDD) \cite{2} is a key technology area in intelligent transportation systems and urban infrastructure maintenance. Through automated detection means, RDD can effectively identify and locate various damages on the road surface, such as cracks, potholes, and cracks. This is essential for the daily maintenance and management of the city, because timely detection and repair of road damage can not only prolong the service life of the road, but also improve the safety of the road and reduce the occurrence of traffic accidents.

Traditional road damage detection mainly relies on manual inspection, this method is not only time-consuming and labor-intensive, but also susceptible to the influence of human factors, resulting in omission or misdetection. With the development of computer vision, machine learning and drone technology \cite{8}, automated road damage detection methods have received more and more attention and research. In particular, image-base and video-based detection techniques \cite{9}, are able to utilize deep learning models such as convolutional neural networks (CNN) \cite{10} , to efficiently and accurately detect and classify road damage \cite{12}.

In recent years, as the application of Unmanned Aerial Vehicles (UAVs) in various fields becomes more and more extensive, road damage detection based on UAVs \cite{13} has gradually become a research hotspot. UAVs have the advantages of flexible flight capability and multi-angle imaging, which can quickly cover a large area of the road area, and significantly improve the detection efficiency. However, due to the diversity and complexity of road damage, such as the irregularity of damage size, shape, and direction, as well as the interference of environmental background, UAV still faces many challenges in the detection process. Therefore, researchers continue to explore new models and algorithms \cite{14} to improve the accuracy and robustness of detection.

Object Detection \cite{15} is one of the core tasks in computer vision, aiming at recognizing all targets in an image or video and determining their locations and categories. With the development of deep learning, especially the application of convolutional neural networks (CNNs), object detection algorithms have made significant progress in accuracy and efficiency. Classical methods such as the R-CNN family \cite{16} achieve accurate detection through region extraction and classification, while single-stage detection algorithms such as YOLO \cite{18} and SSD \cite{19} achieve real-time performance through dense prediction. In recent years, lightweight \cite{20} and Transformer \cite{21} based detection models have further advanced this field, which are widely used in real-world scenarios such as autonomous driving, security surveillance and so on.

Although road damage detection (RDD) has made significant progress in the field of computer vision and deep learning in recent years \cite{25} , the field still faces many challenges, such as the effective extraction and fusion of multi-scale features, the interference of complex backgrounds, and the need for real-time detection. In order to solve these problems, many researchers have proposed different methods and improvements. He et al. \cite{26} proposed a road damage detector based on a local perceptual feature network for the problem of uncertainty in the proportion of the damaged area in road damage detection (RDD). The multiscale feature map extracted by CSP-Darknet53 was used to map it to a local perceptual feature network LFS-Net Multi-scale fusion is performed to generate local feature representations, and finally the feature maps are fed into the detection head for detection, Ning et al. \cite{28} proposed an effective detection method of multiple types of pavement distress based on low -cost front-view video data, modified the YOLOv7 framework, and introduced distributed displacement convolution, efficient aggregation network structure, and improved spatial feature-pattern pyramid structure (SPPCSPD) with similarity-attention mechanism (SimAM) integrated into the model. Wang et al. \cite{35} proposed a social media image dataset for object detection of disaster-induced road damage (SODR) and an integrated learning method based on the attention mechanism of YOLOv5 network. Zhu et al. \cite{38} for the problem of uniformly identifying multiple road daily maintenance inspection proposed a multi-target automatic detection method based on UAV-assisted road daily maintenance inspection, UM-YOLO, which incorporates Efficient Multiscale Attention Module (EMA) in the C2f module in the backbone \cite{41} , as well as Bi-FPN in the neck part, and use lightweight convolution GSConv for convolution operation. Zhang et al. \cite{44} proposed a fast detection algorithm FPDDN for real-time road damage detection, which inherits the deformation transformer that improves irregular defects detection capability, the lightweight D2f module and the SFB downsampling module, which enhances the model's ability of extracting the global damage features, and reduces the loss of small-scale defect information. 

However, these methods still have some shortcomings. For example, multi-scale feature extraction and fusion may be difficult to ensure the stability and accuracy of detection when facing complex backgrounds or irregular damage shapes. In addition, in scenarios with high demand for real-time detection, the computational complexity and inference speed of the model still need to be further optimized to achieve efficient detection on resource-limited devices. To overcome these challenges, this study proposes an RDD model for UVA that flexibly adapts to road damage at multiple scales and automatically removes background interference, called Dynamic Scale-Aware Fusion Detection Model (RT-DSAFDet). Specifically,the contributions of this study are summarized as follows:

1. In this study, we designed a feature extraction module (Flexible Attention, FA module) that can flexibly adapt to changes in the shape and background of road damage, which effectively improves the detection stability and accuracy of the model in complex scenes.

2. A multi-scale sensing and adaptive Shape and background (DSAF module) was designed. By integrating multi-scale features and adapting road damage to different shapes and backgrounds, DSAF module significantly improved the multi-scale feature extraction and fusion capabilities of the model, and further improved the detection performance. By fusing multi-scale features and adapting to different shapes and backgrounds of road damage, the DSAF module significantly enhances the model's capability in multi-scale feature extraction and fusion, which further improves the detection performance.

3. In order to improve the computational efficiency of the model while maintaining the accuracy, an efficient downsampling module (Spatial Downsampling, SD module) is designed in this study, which dramatically reduces the number of parameters and the computational complexity of the model, making it more suitable for real-time detection needs.

4. In this study, a novel road damage detection model ( RT-DSAFDet ) is proposed and designed, and the experimental results on the publicly available datasets UAV-PDD2023 and MS COCO2017 val show that the model outperforms the current state-of-the-art real-time object detection in terms of both accuracy and efficiency. models, demonstrating excellent performance and wide applicability.

\section{Method}
In this section, we provide a comprehensive overview of the proposed model. We provide a detailed description of each module in the network model and clarify their respective functions. First, an explanation of the overall model will be provided, followed by a detailed explanation of the modules involved as well as the structure, including the Flexible Attention (FA) module, the Dynamic Scale-Aware Fusion (DSAF) module, and the Spatial Downsampling (SD) module.

\subsection{Overview}

The structural diagram of the RT-DSAFDet model is divided into three main parts: Backbone, Multi-Scale Fusion, and Head, each of which carries out a different function and together constitute the entire detection system.

\begin{figure}[ht]
\centering
\includegraphics[width=\columnwidth]{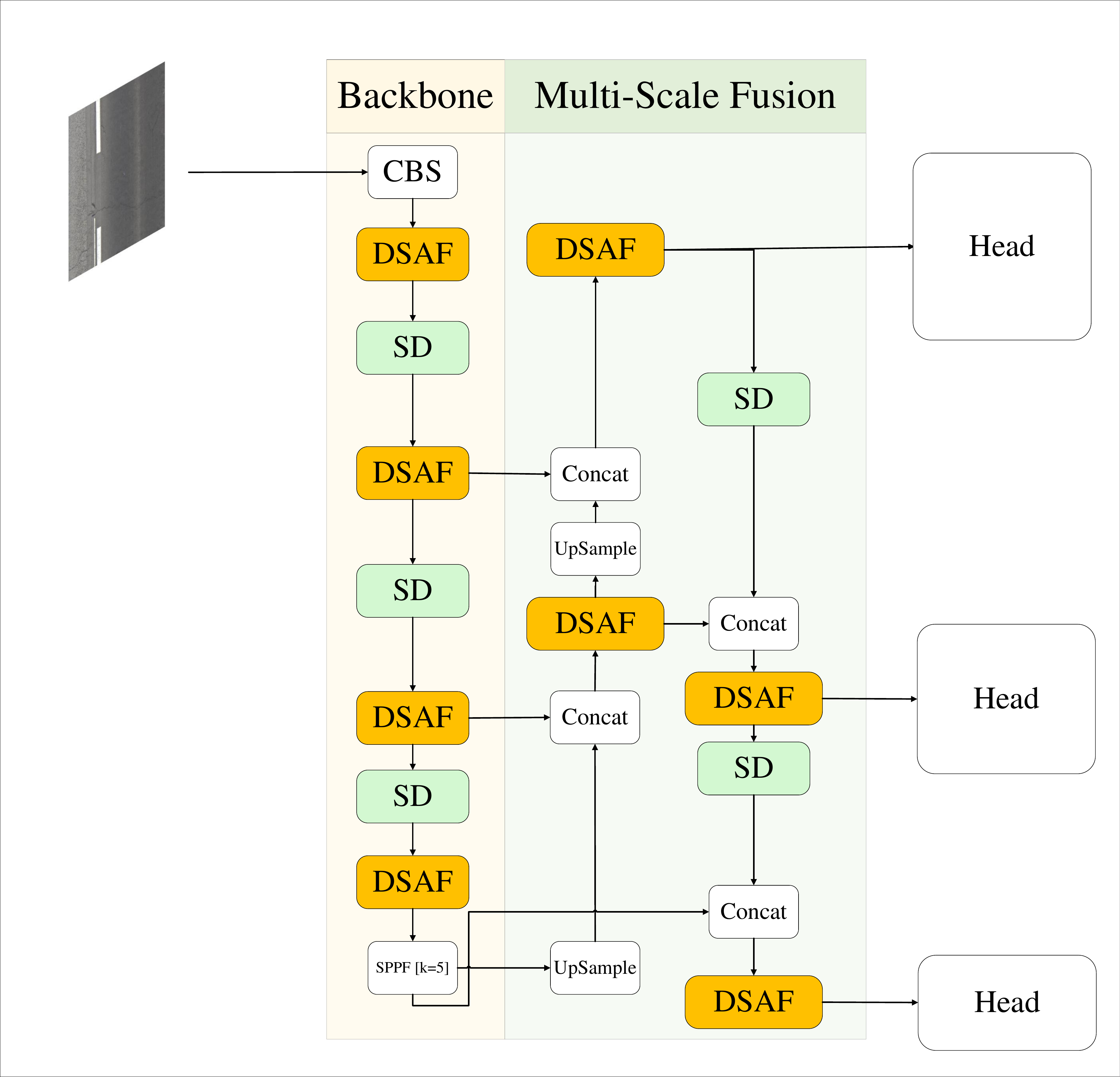} 
\caption{Proposed modeling framework diagram. The RT-DSAFDet model first extracts the image features through the backbone network, and then uses DSAF and up and down sampling to fuse the information of different scales in the multi-scale fusion module. }
\label{fig1}
\end{figure}

In the Backbone section, the model first performs basic feature extraction on the input image through the CBS (Conv-BN-SiLU) module. This process utilizes classical convolutional operations to capture the low-level features of the image. Next, the extracted features are sequentially processed through multiple DSAF and SD modules.The DSAF module focuses on the dynamic fusion of multi-scale features to ensure that the model is able to adapt to road impairments of different sizes and shapes, while the SD module reduces the number of features in the feature map through down sampling operation to reduce the size of the feature map, thus reducing the computational complexity and improving the efficiency of the model.

The core part of the model is Multi-Scale Fusion, whose main task is to fuse feature maps from Backbone at different scales. Through multiple UpSample and Concat operations, this part effectively combines the feature maps at each level, enhancing the model's multi-scale perception capability. In this process, the feature maps are continuously fused and optimized to ensure that the model can acquire rich and consistent feature information in the final detection stage. In order to maintain feature consistency and optimize performance, the Multi-Scale Fusion part again uses the DSAF and SD modules, which allows features to be further enhanced as they are passed and fused.

In the Head section, the model uses multiple detection heads to process the fused feature maps. Each detection head is responsible for different scales of feature maps, which ensures the accuracy and comprehensiveness of the detection results. Specifically, the detection head processes the feature maps through a series of convolutional, classification and regression layers to output information such as the category, location and bounding box of road damage. 

The RT-DSAFDet model forms a powerful road damage detection system by combining preliminary feature extraction with Backbone, multi-scale feature fusion and optimization with Multi-Scale Fusion, and accurate detection with Head part. The system is particularly suitable for road damage detection tasks in complex scenarios, and is able to provide highly accurate detection results while maintaining high efficiency. The experimental results also verify the excellent performance of the model on various public datasets, proving its wide applicability in practical applications.

\subsection{Flexible Attention Module}

This is a structural diagram of the Flexible Attention (FA) module, Fig~\ref{fig3} showing the internal processing flow of the module.The FA module is designed to enhance the model's adaptability to shape and context, especially when dealing with complex road damage, and to allow flexibility in adjusting attention to different features. A detailed description of each part of the module is given below:

\begin{figure}[ht]
\centering
\includegraphics[width=\columnwidth]{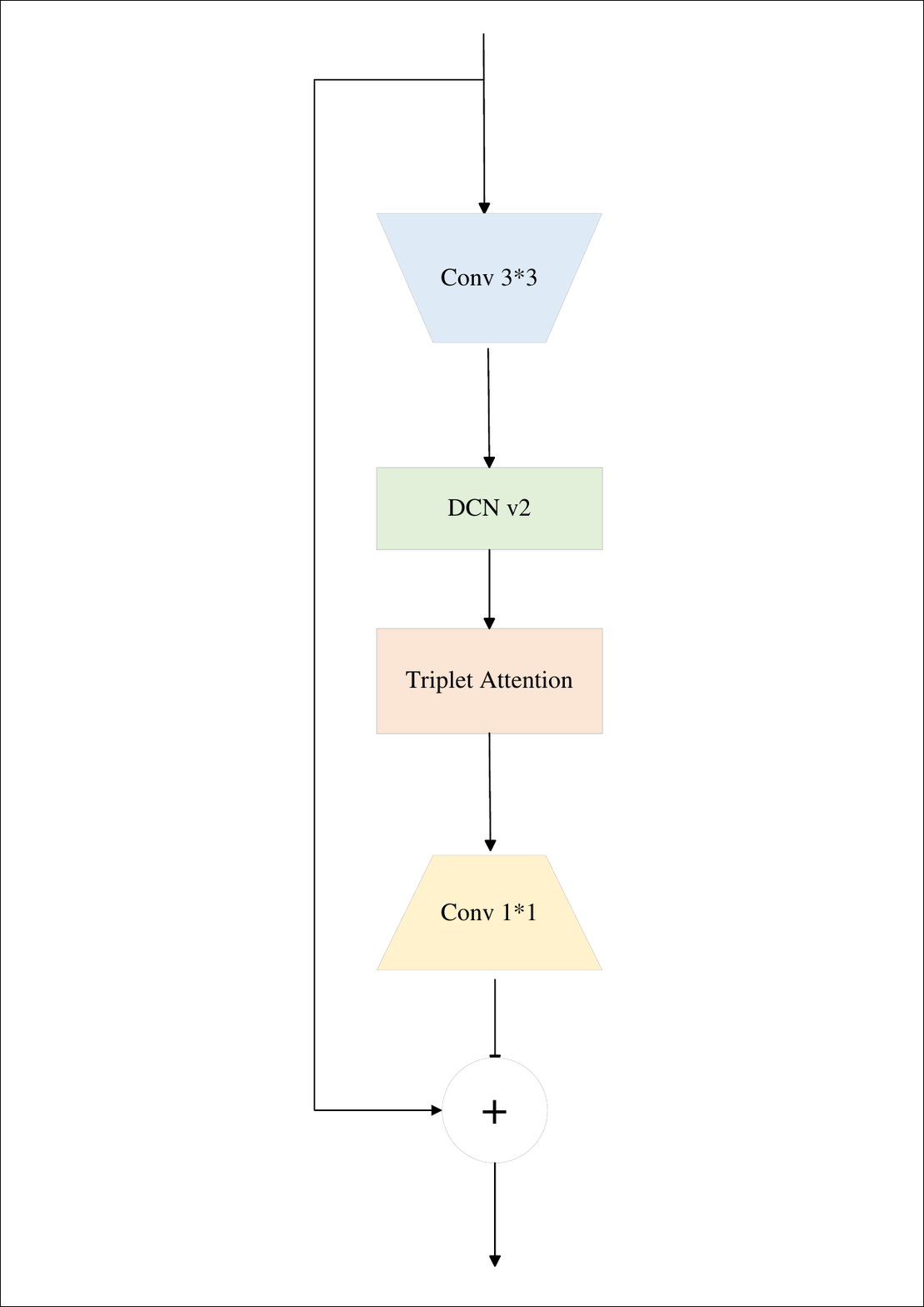}
\caption{Structure of Flexible Attention module. The FA module differs from ordinary convolutional blocks in that it uses deformable convolution (DCNv2) to accommodate irregularly shaped targets and enhances the capture of key features and the suppression of background noise through a triad attention mechanism. }
\label{fig2}
\end{figure}

The Flexible Attention (FA) module is a key component in the RT-DSAFDet model, specifically designed to enhance the model's adaptability when dealing with complex road impairments. The module first performs an initial extraction of the input features through a 3 × 3 convolutional layer to enhance the spatial information. Next, the shape and size of the convolutional kernel are adaptively adjusted using Deformable Convolutional Network v2 (DCNv2) \cite{50}, which enables the model to better capture irregularly shaped and different scales of road damage features. Then, the module introduces the Triple Attention mechanism (Triple Attention) \cite{51}, which calculates the attention weights in the channel, spatial, and orientation dimensions separately, further improving the model's focus on key features and helping to distinguish road damage from background noise. After that, a 1 × 1 convolutional layer is passed to integrate the multidimensional features, which compresses the channel dimension to reduce computational complexity while retaining important feature information. Finally, the FA module sums the original input features with the processed features through a jump join, which preserves the integrity of the input features while enhancing the richness and robustness of the feature representation. This design enables the FA module to provide more accurate and efficient road damage detection capabilities when dealing with complex shapes and background variations.

\subsection{Dynamic Scale-Aware Fusion Module}

The DSAF module (Dynamic Scale-Aware Fusion) is one of the core components in the RT-DSAFDet model, which is designed to significantly improve the model's detection capability in complex road damage scenarios by dynamically fusing multi-scale features. Fig~\ref{fig4} showing the internal processing flow of the module. This module is especially critical for dealing with road injuries with different scales, shapes, and background complexities, as it ensures the accuracy of feature extraction and completeness of details while maintaining high efficiency.

\begin{figure}[ht]
\centering
\includegraphics[width=\columnwidth]{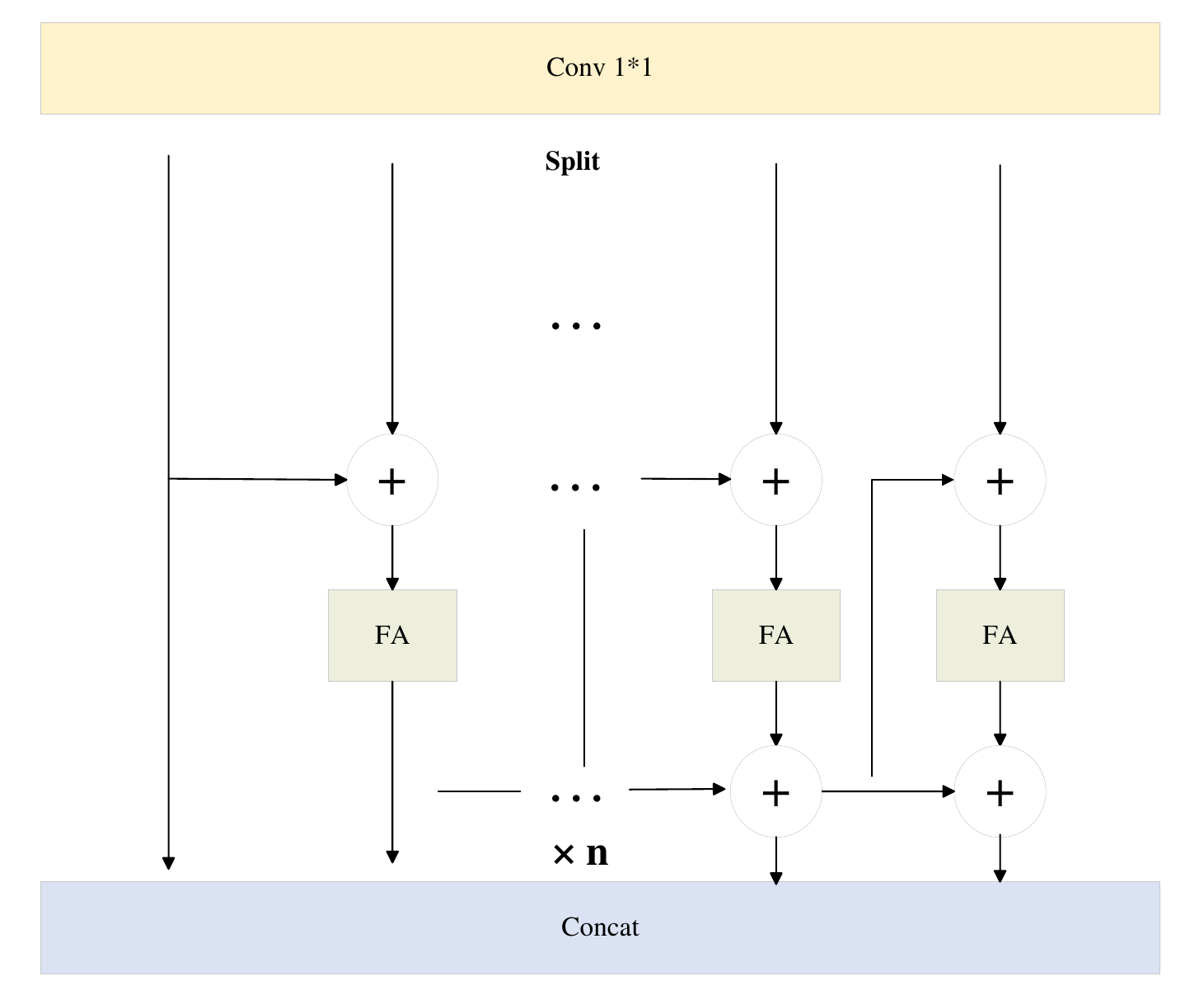}
\caption{Structure of Dynamic Scale-Aware Fusion module. Compared with common feature extraction module, DSAF module introduces multiple FA modules and enhances the ability of multi-scale feature fusion and expression through parallel processing and feature Concat. }
\label{fig3}
\end{figure}

Specifically, the DSAF module first processes the input feature map through a 1×1 convolutional layer, adjusting the number of channels to unify the dimensionality of the input, while effectively reducing the computational complexity and ensuring the efficient use of computational resources. This operation also preserves the key information in the feature map, providing a good foundation for subsequent processing. Next, the processed feature map is partitioned into multiple sub-feature maps. This partitioning allows the module to process features independently at a more detailed scale, which improves the flexibility and accuracy of the model in processing complex scenes. Each sub-feature map is then recursively processed multiple times through the Flexible Attention (FA) module, which plays a crucial role in this process. It is able to adaptively adjust the feature maps at different scales and orientations, resulting in richer and more precise feature representations. Especially when facing road damages with complex shapes and changing backgrounds, Eventually, these sub-feature maps, which have been processed several times, are reintegrated through the concatenation operation (Concat) to form a fused feature map that integrates the detailed information of each sub-feature map. This fused feature map not only retains all the important multi-scale information, but also enhances the overall representation of the features, enabling the model to better identify and localize road damage in subsequent detection tasks. 

\subsection{ Spatial Downsampling Module}

The Spatial Downsampling (SD) module is an important component of the RT-DSAFDet model for efficient spatial downsampling. Its main function is to significantly reduce the size of the feature map through the downsampling operation, thereby reducing the computational complexity, while ensuring that key information is retained to guarantee the performance of the model in subsequent processing.

\begin{figure}[ht]
\centering
\includegraphics[width=\columnwidth]{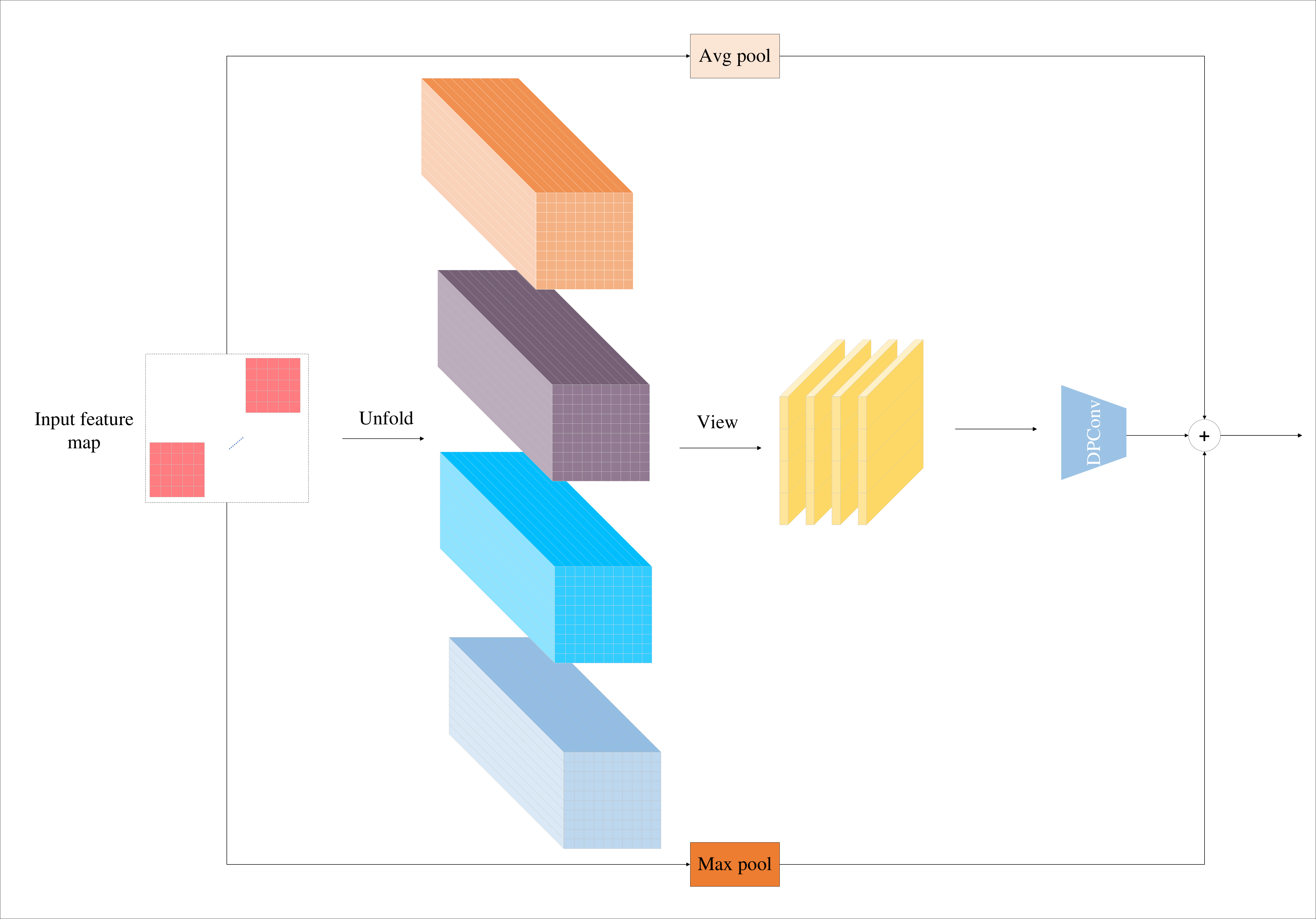}
\caption{Structure of Spatial Downsampling module. Compared with the common Downsampling module, Spatial Downsampling module introduces the combination of maximum pooling and average pooling, so as to retain key information more effectively. }
\label{fig4}
\end{figure}

First, the SD module receives the input feature map and unfolds it into small localized regions through the "Unfold" operation. This unfolding process divides the feature map into smaller units, allowing the downsampling operation to flexibly handle features at different scales. Subsequently, through the "View" operation, the module rearranges these small cells into shapes suitable for downsampling, a process that helps the module to better preserve the key local information when downsizing the feature map. Next, the SD module samples the feature map by two strategies: Max Pool and Avg Pool. Max Pooling selects the maximum value in the local region and focuses on retaining the most salient features, this approach ensures that the model does not lose the critical high response features during the downsampling process; Avg Pooling calculates the average value in the local region, which helps to retain the overall information and smoothes the representation of the feature map. The pooled feature maps are then reintegrated through the "+" operation to form a feature map that incorporates multi-scale information. In the end, the feature map output from the SD module not only significantly reduces the computational effort, but also retains the key information in the input feature map, which provides an efficient and informative input for the subsequent detection module.

The operation of DPConv can be explained by the following formula:

1. Depthwise Convolution: Convolution is performed on each input channel separately.

\begin{equation} 
Y^{(k)}(i,j) = \sum_{m=1}^{M_k}\sum_{n=1}^{N_k}X^{(k)}(i+m,j+n)\cdot K^{(k)}(m,n)
\end{equation}

$\mathbf{X}^{(k)}$ is the feature diagram of the KTH input channel,$\mathbf{K}^{(k)}$ is the convolution kernel corresponding to the KTH channel, $\mathbf{Y}^{(k)}$is the output feature map after Depthwise convolution.

2. Pointwise Convolution Performs 1x1 Convolution on Depthwise Convolution to integrate information on all channels.

\begin{equation}
Z(i,j) = \sum_{k=1}^{K} Y^{(k)}(i,j) \cdot W^{(k)}
\end{equation}

C is the number of input channels,$\mathbf{W}^{(k)}$ is the weight of Pointwise convolution, and Z(i,j) is the output feature map after Pointwise convolution.

\section{Experimental Details}

In this section, a brief overview of the experimental setup and related resources is presented. Next, the experimental dataset, the experimental setup and the evaluation metrics are presented in turn.

\subsection{Datasets}
1) UAV-PDD2023 dataset: The UAV-PDD2023 dataset \cite{52} is a benchmark dataset specifically designed for road damage detection using unmanned aerial vehicles (UAVs). The dataset provides researchers and practitioners in the fields of computer vision and civil engineering with an important resource that can help in developing and evaluating machine learning models for detecting and categorizing various road damages. 

2) MS COCO2017 (Microsoft Common Objects in Context 2017): The MS COCO2017 dataset \cite{53} is one of the most widely used benchmark datasets in the field of computer vision, and is mainly used for the tasks of object detection, image segmentation, keypoint detection and image caption generation. The dataset is published by Microsoft to advance research and development in the field of vision, especially in object recognition and understanding in real-world scenarios.

\begin{table*}[ht]
\centering
\caption{Details of the dataset. } \label{t2}
\begin{tabular}{c | c | c | c | c | c | c }
\toprule
 \multirow{2}{*}{dataset name} & Experimental  & Number of & Number of  &Number& Number of samples  & Test set  \\
 &  image size &  categories &  pictures & of frames&   in the training set & sample size \\
\cmidrule(lr){1-7}
UAV-PDD2023 & \multirow{3}{*}{ 640*640 } & 6 & 2440 & 11158 & 1952 & 488 \\
\cmidrule(lr){1-1} \cmidrule(lr){3-7}
MS COCO2017 &  & 80 & 123287 & 886000 & 118287 & 5000 \\
\bottomrule
\end{tabular}
\end{table*}

\subsection{Experimental Setup}
The experimental program was executed on Windows 11 operating system with NVIDIA GeForce RTX 4090 graphics card driver. The deep learning framework was selected as Pytorch+cu version 11.8 with 2.0.1, the compiler was Jupyter Notebook, Python 3.8 was used as the specified programming language, and all the algorithms used in the comparative analyses were operationally consistent and ran in the same computational settings. The image size was normalized to 640 × 640 × 3. The batch size was 8, the optimizer was SGD, the learning rate was set to 0.001, and the number of training periods was 300.

\subsection{Evaluation Metrics}
In this study four key metrics precision , recall, mAP50 and mAP50-95 \cite{54} were used to evaluate the performance of the detection model. 

\section{Experimental Results and Discussion}

In order to validate the superior performance of the RT-DSAFDet object detection model proposed in this paper, a series of validations are conducted on the above dataset and several evaluation metrics mentioned above are used for evaluation and analysis.

Firstly, this section introduces the current mainstream object detection models and conducts comparison experiments with the model RT-DSAFDet proposed in this paper to prove the superiority of the proposed model. Then, the results of the model proposed in this paper are evaluated, including the analysis of UAV-PDD2023 dataset comparison experimental results, UAV-PDD2023 dataset comparison experimental model recognition results. Finally, the validity of the module as well as the structure designed in this paper is verified by ablation experiments.

\subsection{Comparative experiments }

In order to validate the performance of the proposed model, we compared the RT-DSAFDet trained using the well cover dataset with the YOLOv5 \cite{37}, YOLOv8 \cite{41}, YOLOv9 \cite{55}, YOLOv10 \cite{56}, and RT-DETR \cite{57} models. By this experiments , the superior performance of the model was demonstrated. The mAP50 compared with YOLOv5-m, YOLOv8-m, YOLOv9-t, YOLOv10-m, and RT-DETR-l x0.5 were 12.6\%, 5.8\%, 9.5\%, 11.1\%, and 8.6\% higher, respectively.
\begin{figure*}[ht]
  \centering
  \includegraphics[width=2\columnwidth]{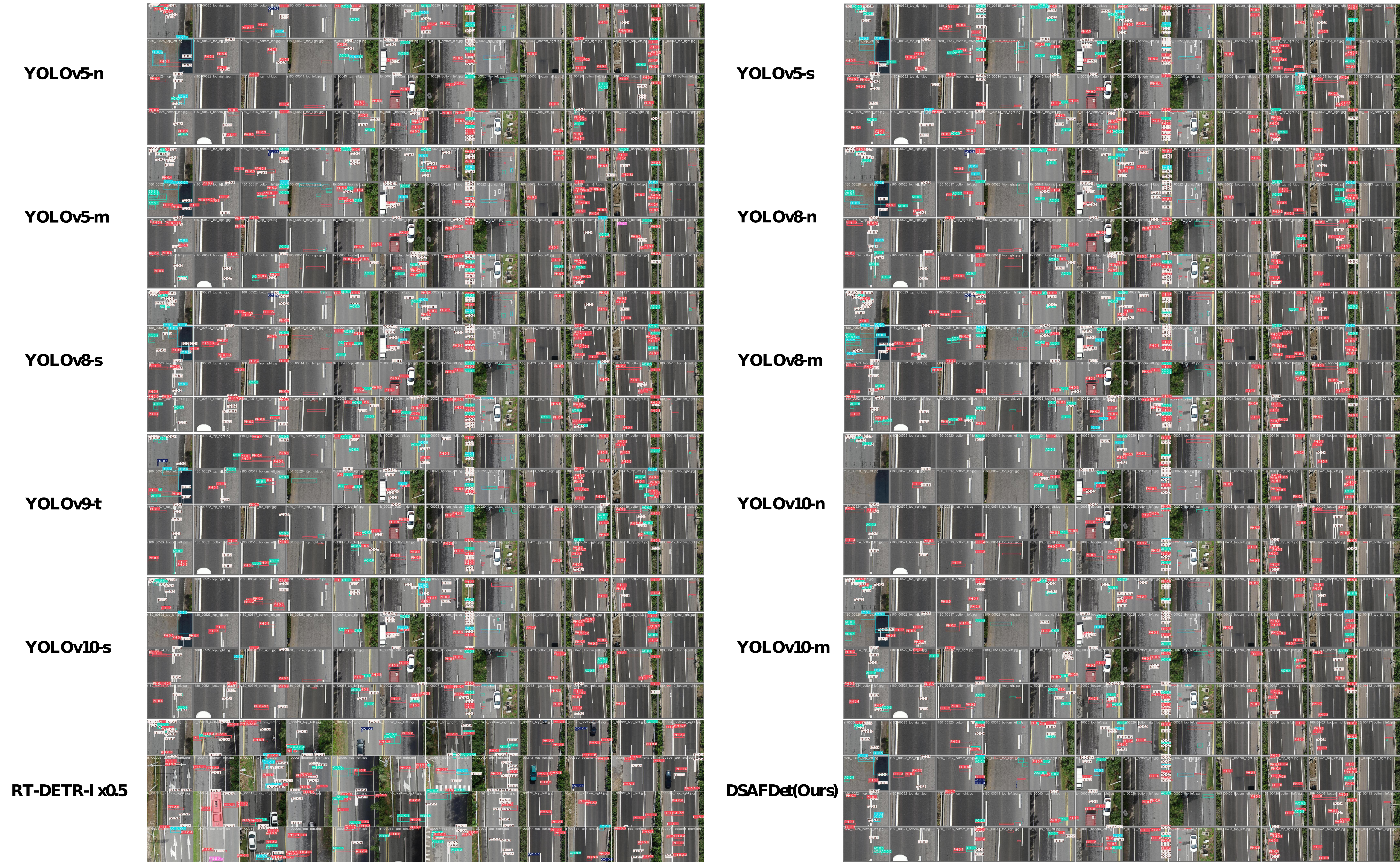}
  \caption{Recognition results of UAV-PDD2023 dataset comparing experimental models. }
  \label{fig5}
  \end{figure*}

\begin{table*}[ht]
\centering
\caption{Comparison experiment results of UAV-PDD2023 dataset} \label{t3}
\begin{tabular}{ p{22mm}  p{5mm}  p{5mm}  p{8mm}  p{10mm}  p{8mm}  p{8mm}  p{12mm} }
\toprule
\multirow{2}{*}{Model} &\multirow{2}{*}{P(\%)}  & \multirow{2}{*}{R(\%)} & mAP  & mAP50  & Params& FLOPs & Model  \\
&  &  &50(\%) &  -95(\%) & (M) & (G) & Size(MB) \\
\hline
YOLOv5-n & 42.5 & 35.4 & 32.3 & 13.9 & 2.5 & 7.1 & 5.3 \\
YOLOv5-s & 52.9 & 40.2 & 40.2 & 18.0 & 9.1 & 23.8 & 18.5 \\
YOLOv5-m & 59.4 & 40.2 & 41.6 & 19.0 & 25.0 & 64.0 & 50.5 \\
YOLOv8-n & 60.9 & 42.2 & 43.6 & 19.5 & 3.0 & 8.1 & 6.3 \\
YOLOv8-s & 49.2 & 41.6 & 41.1 & 18.9 & 11.1 & 28.5 & 22.5 \\
YOLOv8-m & 61.7 & 47.0 & 48.4 & 23.8 & 25.8 & 78.7 & 52.1 \\
YOLOv9-t & 55.9 & 45.7 & 44.7 & 20.9 & 2.0 & 7.6 & 4.7 \\
YOLOv10-n & 42.5 & 35.7 & 31.1 & 14.3 & 2.7 & 8.2 & 5.8 \\
YOLOv10-s & 52.9 & 37.9 & 38.4 & 17.7 & 8.0 & 24.5 & 16.5 \\
YOLOv10-m & 53.8 & 43.0 & 43.1 & 21.4 & 16.5 & 63.5 & 33.5 \\
RT-DETR-l x0.5 & 56.1 & 44.4 & 45.6 & 19.2 & 18.7 & 50.3 & 38.1 \\
\hline
\textbf{\multirow{2}{*}{RT-DSAFDet } }& \textbf{\textbf{68.7}} & \textbf{\textbf{50.5}} & \textbf{\textbf{54.2}} & \textbf{\textbf{27.9}} & \textbf{\textbf{1.8}} & \textbf{\textbf{4.6}} & \textbf{\textbf{4.0}} \\
&\textbf{($\uparrow$7.0)}&\textbf{($\uparrow$3.5)}&\textbf{($\uparrow$6.2)}&\textbf{($\uparrow$4.1)}&\textbf{($\downarrow$0.2)}&\textbf{($\downarrow$3.5)}&\textbf{($\downarrow$0.7)}\\
\bottomrule
\end{tabular}
\end{table*}

1)Analysis of the results of the comparison experiments on the UAV-PDD2023 dataset: As shown in Table1. RT-DSAFDet has a precision (P) of 68.7\% and a recall (R) of 50.5\%, which are the highest metrics among all the compared models. 

\begin{table*}
  \centering
  \caption{ Results of comparison experiments on MS COCO2017 dataset. } \label{t5}
  \begin{tabular}{ c | c | c | c | c | c | c }
  \toprule
  Model & Resolution & mAP50 & mAP50-95 & Params & FLOPs& Model Size \\
  \cmidrule(lr){1-7}
  YOLOv5-n & 640*640 & --  & 28.4 & 1.9 & \textbf{\textbf{4.5}} & \textbf{\textbf{3.9}} \\
  \cmidrule(lr){1-7}
  YOLOv8-n & 640*640 & --  & 37.3 & 3.2 & 8.7 & 6.2 \\
  \cmidrule(lr){1-7}
  YOLOv9-t & 640*640 & 53.1 & 38.3 & 2.0 & 7.7 & 4.4 \\
  \cmidrule(lr){1-7}
  YOLOv10-n & 640*640 &  -- & \textbf{\textbf{38.5}} & 2.3 & 6.7 & 5.7 \\
  \cmidrule(lr){1-7}
  \textbf{\textbf{RT-DSAFDet }}& 640*640 & \textbf{\textbf{53.6}} \textbf{($\uparrow$0.5})& 38.3\textbf{($\downarrow$0.2)} & \textbf{\textbf{1.8}}\textbf{($\downarrow$0.1)} & 4.6\textbf{($\uparrow$0.1)} & 4.0\textbf{($\uparrow$0.1)} \\
  \bottomrule
  
  \end{tabular}
  
  \end{table*}

The precision metric reflects the accuracy of the model when the prediction is a positive sample, while the recall reflects the model's ability to capture all true positive samples.RT-DSAFDet's leadership in both metrics demonstrates its ability to not only accurately identify damage, but also capture as much of the damage as possible when dealing with complex road damage scenarios where all damage is present. This is particularly important for road maintenance and safety monitoring, ensuring that potential problems are fully detected.RT-DSAFDet achieves 54.2\% on the mAP50 metric, which means that the model's average detection accuracy is extremely high and significantly outperforms that of other models under looser IoU (intersection and concurrency ratio) thresholds. For example, compared to YOLOv5-n (32.3\%) and YOLOv10-n (31.1\%), RT-DSAFDet's mAP50 is 21.9\% and 23.1\% higher, respectively. More notably, RT-DSAFDet also achieves a high score of 27.9\% on the more stringent mAP50-95 metric, which indicates that it maintains high detection performance under different IoU thresholds, demonstrating robustness and adaptability to various complex scenarios. This is especially important when dealing with road damage detection, as different types of damage may exhibit different characteristics under different IoU thresholds. In terms of computational efficiency, RT-DSAFDet also performs well. It has only 1.8M parameters, 4.6G FLOPs, and a model size of only 4.0MB.These values indicate that RT-DSAFDet significantly reduces the computational resource consumption while maintaining high accuracy. In contrast, YOLOv8-m has 25.8M parameters, 78.7G FLOPs, and 52.1MB model size, which are much higher than RT-DSAFDet, indicating that our model is more lightweight and suitable for running in resource-constrained environments, such as embedded systems or mobile devices. This efficient design makes RT-DSAFDet not only perform well in laboratory environments, but also provides the possibility of efficient deployment in real applications.

2)Analysis of the recognition results of the comparative models on the UAV-PDD2023 dataset: As shown in Fig~\ref{fig5}.YOLOv5-n, YOLOv5-s, and YOLOv5-m show some differences in the recognition process.YOLOv5-n is relatively lightweight, and thus lacks in detection accuracy, especially in the recognition of some minor damages (e.g., small cracks, minor pits) prone to missed or false detections.The performance of YOLOv5-s and YOLOv5-m is improved, especially in the sense that more features are correctly detected while the number of false detections is reduced. However, these models still show shortcomings when dealing with complex scenarios, such as the tendency of label confusion in the presence of multiple road markings or interference from mixed objects.The YOLOv8 series of models have improved to a certain extent with respect to YOLOv5.YOLOv8-n and YOLOv8-s show improvements in detection accuracy, especially in distinguishing between different types of damages (e.g., longitudinal cracks and transverse cracks) with more accurate performance. However, YOLOv8-n and YOLOv8-s still have some under-detection in some complex scenarios, while YOLOv8-m is the best in terms of detection stability and accuracy, and is able to accurately recognize most of the road damages, but there may still be a few mis-detections in very small damages or complex backgrounds. The YOLOv9-t model has improved the detection accuracy, and can identify various types of road damage well, especially in the case of damage with low contrast, it still maintains a high detection rate. YOLOv10-n and YOLOv10-s in the YOLOv10 series of models show strong adaptability in some test scenarios, especially when dealing with multi-object dense scenes perform more stably, but the overall precision and recall are still slightly lower than that of the YOLOv9-t. The performance of the RT-DETR-l x0.5 model is somewhat improved compared to the YOLO series, especially the stronger detection ability in complex backgrounds. This can be attributed to the model's stronger feature extraction capability and the fusion processing of multi-scale features. Nevertheless, in some very small or low contrast damage detection, RT-DETR-l x0.5 still has a small amount of missed detection.The RT-DSAFDet model performs well in all test scenarios. Compared with other models, the model can more accurately detect various types of road damage, including some small cracks and minor road breaks, etc. RT-DSAFDet, thanks to its multi-scale feature fusion and adaptive attention mechanism, can not only maintain high detection accuracy in complex backgrounds, but also effectively avoid leakage and misdetection, especially in multi-object dense and complex scenes, which is particularly The performance is especially outstanding in multi-object dense and complex scenes.

\begin{table*}[ht]
  \centering
  \caption{ Results of ablation experiments on UAV-PDD2023 dataset. } \label{t6}
  \begin{tabular}{ l | l | l | l | l | l | l | l | l }
  \toprule
  DASF & SD & P & R & mAP50 & mAP50-95 & Params& FLOPs& Model Size\\
  \cmidrule(lr){1-9}
    &   & 60.9 & 42.2 & 43.6 & 19.5 & 3.0 & 8.1 & 6.3 \\
  \cmidrule(lr){1-9}
  $\checkmark$ &   & 66.3\textbf{($\uparrow$5.4)} & \textbf{\textbf{54.9}}\textbf{($\uparrow$12.7)} & \textbf{\textbf{58.3}}\textbf{($\uparrow$14.7)} & \textbf{\textbf{30.0}}\textbf{($\uparrow$10.5)} & 2.65\textbf{($\downarrow$0.35)} & 7.4\textbf{($\downarrow$0.7)} & 5.7\textbf{($\downarrow$0.6)} \\
  \cmidrule(lr){1-9}
    & $\checkmark$ & 59.6\textbf{($\downarrow$1.3)} & 46.7\textbf{($\uparrow$4.5)} & 51.0\textbf{($\uparrow$6.4)} & 25.0\textbf{($\uparrow$5.5)} & 2.7\textbf{($\downarrow$0.3)} & 7.5\textbf{($\downarrow$0.6)} & 5.7\textbf{($\downarrow$0.6)} \\
  \cmidrule(lr){1-9}
  $\checkmark$ & $\checkmark$ & \textbf{\textbf{68.7}}\textbf{($\uparrow$7.8)} & 50.5\textbf{($\uparrow$8.3)} & 54.2 \textbf{($\uparrow$10.6)}& 27.9\textbf{($\uparrow$8.4)} & \textbf{\textbf{1.8}}\textbf{($\downarrow$1.2)} & \textbf{\textbf{4.6}}\textbf{($\downarrow$3.5)} & \textbf{\textbf{4.0}}\textbf{($\downarrow$2.3)} \\
  \bottomrule
  \end{tabular}
  \end{table*}
\subsection{Generic Object Detection Experiments }

Experimental results on the MS COCO2017 dataset show that the RT-DSAFDet model performs well in several key performance metrics, in particular, it achieves detection accuracy comparable to that of YOLOv9-t while maintaining a low computational resource consumption. Specifically, RT-DSAFDet achieves 53.6\% for mAPval 50 and 38.3\% for mAPval 50-95, both of which are on par with the performance of YOLOv9-t, showing the model's detection stability and accuracy under different IoU thresholds.

In addition, the computational efficiency of RT-DSAFDet is significantly better than other models. Its number of parameters is only 1.8M, FLOPs is 4.6G, and model size is 4.0MB, all of which indicate that RT-DSAFDet achieves a high degree of compactness and efficiency in its design. Despite the relatively small model size, RT-DSAFDet is still able to lead in detection accuracy, making it ideal for applications in resource-constrained environments such as mobile devices or embedded systems.

\subsection{Ablation Experiments}
In this ablation experiment, YOLOv8-n was used for the benchmark model, and the performance indicators of the model were analyzed in detail by introducing DASF and SD modules.

\begin{figure}[t]
\centering
\includegraphics[width=\columnwidth]{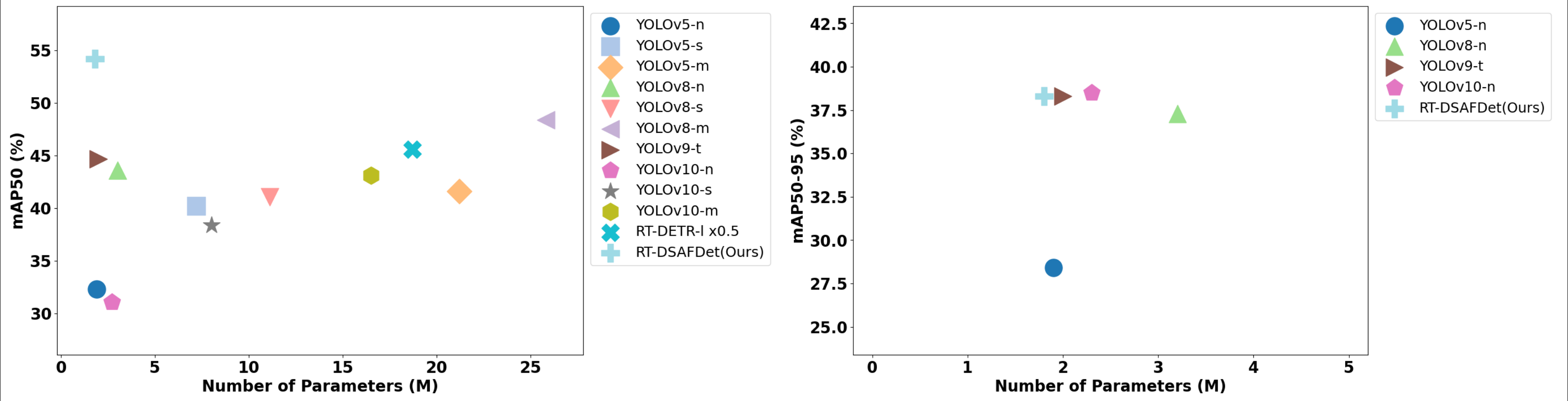}
\caption{Summary of UAV-PDD2023 and MS COCO2017 datasets mAP and Parameters. }
\label{fig6}
\end{figure}

The benchmark model has a precision (P) of 60.9\%, a recall (R) of 42.2\%, a mAP50 of 43.6\%, and a mAP50-95 of 19.5\% when no additional modules are introduced. Although this benchmark model already has some detection capability, it shows obvious limitations when dealing with complex multi-scale features and complex background scenes, resulting in less than ideal detection precision and recall. With the introduction of the DASF (Dynamic Scale-Aware Fusion) module, the performance of the model is significantly improved. Specifically, the precision is increased to 66.3\%, the recall is improved to 54.9\%, the mAP50 rises to 58.3\%, and the mAP50-95 reaches 30.0\%. the DASF module effectively enhances the model's capability of capturing and fusing the information of different scales in the complex scene through the dynamic fusion of multiscale features, and significantly improves the accuracy and robustness of the detection. In the case where only the SD module is introduced, the model also shows significant improvement in some of the metrics. Although the precision slightly decreases to 59.6\%, the recall improves to 46.7\%, and the mAP50 and mAP50-95 improve to 51.0\% and 25.0\%, respectively. the SD module reduces the computational complexity through efficient spatial downsampling operations and optimizes the model's performance in large-scale feature processing while maintaining critical information. When both DASF and SD modules are introduced, the model achieves the best performance indicators. The precision is improved to 68.7\%, the recall is 50.5\%, and the mAP50 reaches 54.2\% and mAP50-95 27.9\%. In addition, the number of covariates of the model is significantly reduced to 1.8 M, FLOPs are reduced to 4.6 G, and the model size is reduced to 4.0 MB.These results indicate that the combination of DASF and SD modules greatly optimizes the computational efficiency of the model while improving the detection performance, making it suitable for application in resource-constrained environments while maintaining high precision.

\section{Conclusion}

In this paper, we propose and validate the superior performance of the RT-DSAFDet model in road damage detection tasks by comparing and analyzing several advanced object detection models. A series of experiments, especially on the MS COCO2017 dataset, show that RT-DSAFDet meets or exceeds the current state-of-the-art models (e.g., YOLOv9-t) in key metrics, such as mAP50 and mAP 50-95, while significantly reducing the number of parameters and computational complexity of the model. Although the RT-DSAFDet model proposed in this paper demonstrates excellent performance in road damage detection tasks, it still has some shortcomings. First, although the model achieves a balance in terms of accuracy and efficiency, there is still a possibility of missed or false detections in extremely complex scenarios. This is mainly due to the fact that the current model still has some limitations when dealing with features in extremely small scales or extremely complex backgrounds. In addition, while RT-DSAFDet improves the detection performance, it may need to be further optimized in scenarios with very high real-time requirements (e.g., real-time detection in video streaming) to ensure stable detection performance even at higher frame rates. Our future work will try to design new feature extraction and fusion techniques, especially for the detection of extreme small-scale damage, to enhance the model's ability to perceive features at different scales.

\bibliography{sn-bibliography}

\end{document}